\pdfoutput=1

\documentclass[11pt]{article}

\usepackage{authblk}
\usepackage{acl}
\usepackage{svg}
\usepackage{times}
\usepackage{latexsym}
\usepackage{amsmath}
\usepackage{amssymb}
\usepackage{caption}
\usepackage{verbatim}
\usepackage{subcaption}
\usepackage[T1]{fontenc}

\usepackage[utf8]{inputenc}

\usepackage{microtype}
\author[1]{Debayan Banerjee $^\dag$}
\author[2]{Pranav Ajit Nair $^\dag$}
\author[1]{Ricardo Usbeck}
\author[1]{Chris Biemann}
\affil[1]{Universit\"at Hamburg, Hamburg, Germany}
\affil[1]{\texttt{\{firstname.lastname\}@uni-hamburg.de}}
\affil[2]{Indian Institute of Technology (BHU), Varanasi, India}
\affil[2]{\texttt{pranavajitnair.cse18@itbhu.ac.in}}

\title{The Role of Output Vocabulary in T2T LMs \\ for SPARQL Semantic Parsing}

\begin{document}

\maketitle
\def\thefootnote{\dag}\footnotetext{The authors contributed equally to this work}\def\thefootnote{\arabic{footnote}}
\begin{abstract}
In this work, we analyse the role of output vocabulary for text-to-text (T2T) models on the task of SPARQL semantic parsing. We perform experiments within the the context of knowledge graph question answering (KGQA), where the task is to convert questions in natural language to the SPARQL query language.  We observe that the query vocabulary is distinct from human vocabulary. Language Models (LMs) are pre-dominantly trained for human language tasks, and hence, if the  query vocabulary is replaced with a vocabulary more attuned to the LM tokenizer, the performance of models may improve. We carry out carefully selected vocabulary substitutions on the queries and find absolute gains in the range of 17\% on the GrailQA dataset.

\end{abstract}

\section{Introduction}
\label{intro}

Knowledge Graph Question Answering (KGQA) is the task of finding answers to questions posed in natural language, using triples present in a KG. Typically the following steps are followed in KGQA: 1) Objects of interest in the natural language question are detected and linked to the KG in a step called entity linking.  2) The relation between the objects is discovered and linked to the KG in a step called relation linking. 3) A formal query, usually SPARQL\footnote{\url{https://www.w3.org/TR/rdf-sparql-query/}}, is formed with the linked entities and relations. The query is executed on the KG to fetch the answer. 

Our focus in this work is the query building phase, henceforth referred to as KGQA semantic parsing. The motivation of our work stems from \citet{banerjee}, where minor vocabulary substitutions to handle non-printable special characters for T5 \cite{t5} produced better results on the task of SPARQL semantic parsing. In this work, we extend the idea and replace the entire SPARQL vocabulary with alternate vocabularies.

As in \citet{banerjee}, we replace certain special characters in the SPARQL vocabulary, such as \{ , \} with textual identifiers, as T5 is known to have problems dealing with these special characters \cite{banerjee}. We call this a masked query, and in this work, we test the ability of the models to generate this masked query, given the natural language question as input.

A sample question, the original SPARQL query, and the corresponding masked query are as shown below (for the Wikidata KG \cite{vrandevcic2014wikidata}) :

\textit{Is it true that an Olympic-size swimming pool's operating temperature is equal to \textbf{22.4} ? }  

\begin{verbatim}
ASK WHERE 
{ 
  wd:Q2084454 wdt:P5066 ?obj 
  filter(?obj = 22.4) 
}
\end{verbatim}

\begin{verbatim}
ASK WHERE 
OB
  ent0 rel0 ?obj 
  filter ( ?obj = 22.4 ) 
CB
\end{verbatim}

In the era of pre-trained Language Models (LMs) \cite{devlin,t5} it is common practice to fine-tune models on custom downstream datasets. 
This requires supervised training which results in modification of weights of the models using some training algorithm. 
More recently, the technique of prompting of language models \cite{brown, shin} has been developed, which elicits the desired response from a LM through a task description and a few input-output examples. \citet{brown} shows that such a strategy works better for larger models. It has however been observed that prompt design is brittle in behaviour and displays sensitivity to the exact phrase \cite{shin}. \\
A more recent innovation is that of prompt tuning \cite{lester}, where the task-specific prompt is learnt on a  smaller external neural network. The gradients are computed and flow through the LM, but leave the weights of the LM itself unchanged. Instead, the weights of the prompt tuning network change and produce a custom and continuous prompt which produces the desirable response from the LM.

A similar method is prefix tuning \cite{prefix}, which is known to perform better for generation tasks \cite{https://doi.org/10.48550/arxiv.2210.04457}. In this method, the original inputs and outputs are kept the same, but the input is pre-pended with a continuous prefix learnt in the external network. This prefix allows the model to understand the exact task to be performed by it.

As primary contribution, in this work, we perform an analysis of how the complexity of output vocabularies affects the performance on the KGQA semantic parsing task for prefix and fine-tuned language models. Code and data can be found at \url{https://github.com/debayan/sparql-vocab-substitution}.
    
\section{Related Work}

A study of low-resource semantic parsing using prompt tuning was performed by \citet{schucher-etal-2022-power} on the Top v2 \cite{top} and Overnight \cite{wang-etal-2015-building} datasets. Prompt tuning, while not the same as prefix tuning, still keeps the LM weights frozen while the prompts are learnt on an external network. In their experiments, they perform a single kind of vocabulary substitution but find no noticeable performance improvements. No specific study is made of the change in performance with vocabularies of varying complexities, which is a task we undertake. Another difference is that we perform experiments in the high-resource use case as opposed to low-resource.

Another work which is similar to ours is \citet{unfreeze}, where the authors experiment with prefix tuning on the task of semantic parsing, and find problems with non-standard vocabularies of logical forms. In their case, they work with the TOP v2 \cite{top} and PIZZA \cite{pizza} datasets. The keywords in those datasets consist of words joined by underscores (eg: IN:GET\_REMINDER\_DATA\_TIME ), which poses a problem for the sub-word tokenizer of the transformer based models. They find that fine tuning a model on these datasets outperforms prefix-tuning by a large margin. However, when they add the non-standard keywords to the tokenizer vocabulary and re-train the tokenizer to generate new embeddings for these keywords, fine tuning and prefix tuning perform at par. Our work is different in a few respects: firstly, due to the specific research focus of our group, we experiment with a semantic parsing dataset for KGQA, namely GrailQA \cite{grailqa}.  Secondly, instead of retraining the tokenizer, we perform a simpler procedure of pre-processing the dataset by replacing the current vocabulary with a new vocabulary. We then train the models on this modified dataset, and as a post-processing step, substitute back the original vocabulary in place of the new vocabulary.

\begin{table*}[]
  \centering
  
  \begin{tabular}{|c|c|c|c|c|c|c|c|c|c|c|c|c|}
    \hline
       &\multicolumn{6}{c|}{GrailQA} \\
      \hline
      & \multicolumn{2}{c|}{T5-Small}&\multicolumn{2}{c|}{T5-Base}&& \\
      \hline
      &PT&FT&PT&FT&TSVS&ALFL\\
      \hline
    
    \hline
     char8              & 74.03 & 86.57 & 82.65 & 86.72 &306& 263 \\
     char4             & 76.43& 87.09& 84.92& 87.10 & 159&141\\
     char2             & 83.29& 91.49& 89.83& 92.30 & 90 & 87 \\
     char1            & \textbf{84.89} & \textbf{92.13} &  \textbf{91.24} & \textbf{92.61} & 57 & 57\\
     dictionary        & 82.57& 91.95& 90.93& 92.48& 49& 44\\
     original          & 67.10 & 74.08 & 73.06& 74.45 & 124 & 125\\

    \hline
  \end{tabular}
  
  \caption{Exact match percentages for generated masked SPARQL queries. Best performance is always found in substituted vocabularies. For \textbf{char} settings, accuracy drops as vocabulary and query lengths increase. TSVS = Tokenizer specific vocabulary size, ALFL = Average logical form length, PT = Prefix Tuning, FT = Fine Tuning }
  \label{table1}
\end{table*}

\section{Prefix Tuning}
Prefix tuning prepends a set of tunable weights to every key-value pair in the transformer attention. The transformer attention is represented as follows:
\begin{equation}
     \text{attn}(Q, K, V) = \text{softmax} (\frac{Q\cdot K^{\top}}{\sqrt{d}})V
\end{equation}
where the query $Q$, key $K$ and value $V$ are obtained through affine transformations on the input. $d$ represents the model dimension. Prefix tuning modifies the transformer attention by adding tunable prefixes to $K$ and $V$, thereby modifying $K$ as $K^{\prime} = [h_K;K]$ and $V$ as $V^{\prime} = [h_V;V]$. Here $h_K$ and $h_V$ represent the key prefix and the value prefix respectively.

Following \citet{prefix} we model these prefixes using a two layer MLP as follows:
\begin{equation}
\label{prefix eq}
    \begin{aligned}
       h_K=W_{K,2}f(W_{K,1} E+b_{K,1})+b_{K,2} \\
       h_V=W_{V,2}f(W_{V,1} E+b_{V,1})+b_{V,2}
    \end{aligned}
\end{equation}
where $W \in  \mathbb{R}^{d\times d}$ and $b \in \mathbb{R}^{d}$  are trainable weights and biases respectively. $E \in \mathbb{R}^{C \times d}$ is a trainable embedding matrix with $C$ as the prefix length.

\section{Models and Experimental Setup}

We carry out prefix-tuning and fine-tuning experiments with two versions of the T5 model: namely T5-Small (60 million parameters) and T5-Base (220 million parameters). Questions are fed as input during training while masked SPARQL queries, as described in Section \ref{intro}, are provided as labels for supervision.

For evaluation, we use the exact-match metric. A generated query is matched token by token, while ignoring white-spaces, to the gold query. The percentage of queries matched is reported.

\subsection{Hyper-parameters and Implementation Details}
Throughout our experiments, the prefix length is fixed to $50$. For prefix tuning experiments we use the Adafactor \cite{shazeer2018adafactor} optimizer with a constant learning rate of $0.001$. Fine-tuning experiments are optimized through AdamW \citep{DBLP:conf/iclr/LoshchilovH19} with a square root decay schedule, a maximum learning rate of $0.0015$ and a linear warm-up of $5000$ steps. Our code is implemented with HuggingFace Transformers\footnote{\url{https://github.com/huggingface/transformers}} \cite{wolf-etal-2020-transformers} and OpenPrompt\footnote{\url{https://github.com/thunlp/OpenPrompt}} \citep{DBLP:conf/acl/DingHZCLZS22}. T5-Small experiments were run on 12GB Nvidia GTX-1080 and RTX-2080 GPUs, and T5-Base experiments were run on 48GB Nvidia RTX-A6000.
For fine-tuning, we run each training thrice with three separate seeds for 120 epochs each. For prompt tuning we do the same for 400 epochs. We report the inference results of these trained models on the test sets of the respective datasets.

\section{Vocabulary}

The original vocabulary of the GrailQA dataset consists of 48 words. The T5 tokenizer splits these words into 124 sub-words. This tokenizer specific vocabulary size (TSVS) is seen in the last column of Table \ref{table1}. In the next column, the original average logical form (SPARQL query) length can be seen as 125 tokenized sub-words.

We wish to see how a new output vocabulary affects performance, and as a result, we construct a set of special vocabularies and substitute them in-place of the original SPARQL vocabulary. With reference to the settings in Table \ref{table1}, each vocabulary is as described below:

\textbf{original} The masked SPARQL queries remain as they are. No replacement of the original SPARQL keywords is made with an alternate vocabulary. 

\textbf{dictionary} The SPARQL keywords are replaced with a vocabulary of English words. For example, SELECT may be replaced with DOG, [ may be replaced with CAT etc. During the pre-training phase a LM is likely to have seen such words far more frequently than the SPARQL keywords. This mode tests how the model behaves when the output vocabulary is comprised of well known English words. 

\textbf{char1} The SPARQL keywords are replaced with a single character of the English alphabet, for example, SELECT is replaced with A, WHERE is replaced with B. Additionally, numerical digits from 1-9 are used, and if the size of vocabulary demands more, we add single length special characters, such as * and \$.

\textbf{char2}, \textbf{char4} and \textbf{char8} settings apply vocabulary substitution of 2, 4 and 8 character lengths chosen randomly, constituted from the characters A-Z and digits 0-9. For example, a typical \textbf{char8} substitution would be SELECT  replaced by ATYZGFSD. This setting is designed to test the behaviour of the models when asked to produce more number of tokens per original-vocabulary word. 
A sample of a question, the SPARQL and the corresponding substitutions is provided in the Appendix in Table \ref{tablevocabs}.
\section{Datasets}

For our experiments, we require a dataset which contains a mapping of natural language questions to their corresponding logical forms and is large in size, since we test the high resource use-case.

\textbf{GrailQA} \footnote{\url{https://dki-lab.github.io/GrailQA/}} is based on the Freebase knowledge graph \cite{freebase} and consists of 64,331 questions designed to test three levels of generalisation, ie, i.i.d, compositional and zero-shot. For our purposes, we split the train set itself to three parts, since we are not interested in testing compositional generalisation aspects of the test set of this dataset. We are left with the following configuration:  test: 8868, dev: 4434, train: 31035.

\begin{figure}[htb]
     \centering
     \begin{subfigure}[b]{0.45\textwidth}
         \centering
         \includegraphics[width=\textwidth]{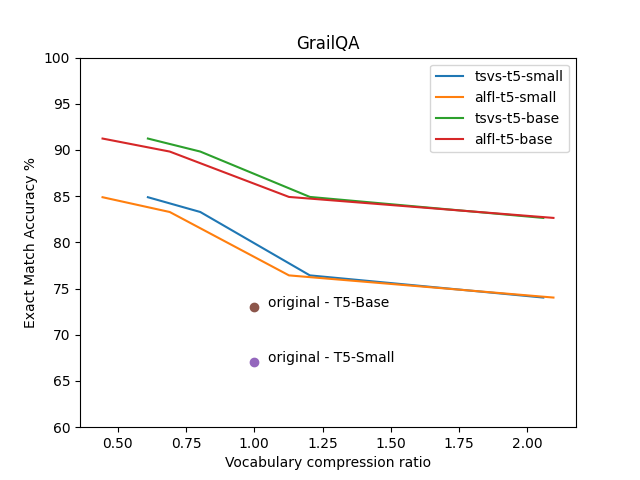}
         \caption{}
         \label{fig:y equals x}
     \end{subfigure}
     \caption{Prefix tuning accuracy drops as vocabulary and query lengths increase for \textbf{char} settings. TSVS = Tokenizer specific vocabulary size, ALFL = Average logical form length}
     \label{ptplot}
     \vspace{-5mm}
\end{figure}

\begin{figure}[htb]
     \centering
     \begin{subfigure}[b]{0.45\textwidth}
         \centering
         \includegraphics[width=\textwidth]{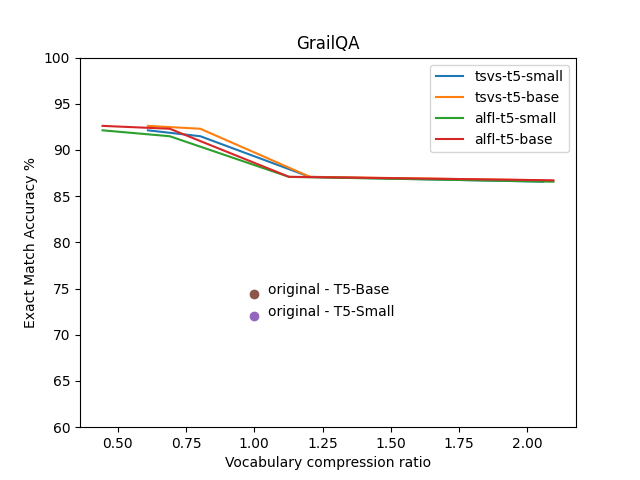}
         \caption{}
         \label{fig:y equals x}
     \end{subfigure}
     \caption{Fine-tuning accuracy drop is more gradual when compared to prefix tuning, and the performance of T5-Small and T5-Base are similar.  TSVS = Tokenizer specific vocabulary size, ALFL = Average logical form length}
     \label{ftplot}
     \vspace{-5mm}
\end{figure}

 

\section{Analysis}

As seen in Table \ref{table1}, the best performance for prefix and fine tuning is achieved for substituted vocabularies. The original vocabulary lags behind in general, which points to the finding, that the choice of an appropriate vocabulary improves performance for semantic parsing. Further, among the substituted vocabularies, the setting \textbf{char8} performs the worst, which signifies the adverse role of the extra decoding load of this vocabulary on the performance of the model. 


This finding is different from that of \citet{schucher-etal-2022-power}, who find their \texttt{in-vocab} setting performing no better overall. They attribute it to the substitutions possibly masking the meanings of the intents, for their given dataset. On the contrary, we find significant gains for GrailQA. It must be noted however, that we perform high-resource prefix tuning while they perform low-resource prompt tuning, and hence results may differ.

As seen in Figure \ref{ptplot}, for the \textbf{char} settings, as the size of vocabulary increases, the prefix tuning accuracy drops. In the said figure, we define vocabulary compression ratio as the size of the new vocabulary divided by the size of the original vocabulary. Apart from vocabulary size, the query length also matters. We dual-define vocabulary compression ratio as the size of query length after substitution of new vocabulary divided by size of original query length, and plot on the same graph. 

When compared to the fine-tuning plot (Figure \ref{ftplot}), prefix tuning has a steeper drop in accuracy, and the performance for T5-Small and T5-Base vary more significantly. It leads to the finding that fine-tuning is less sensitive to vocabulary changes, and the difference in model sizes between T5-Small and T5-Base also seems to matter less. 

In Figures \ref{ptplot} and \ref{ftplot}, it can be seen that the \textbf{original} setting for the masked SPARQL vocabularies produce accuracies which are below the \textbf{char} family vocabulary curves. It suggests that vocabulary compression ratio alone is not a deciding factor in accuracy. If the vocabulary family changes from SPARQL to characters, there is an initial shift in accuracy, and after that the complexity of the character vocabulary further affects the accuracy.

In Table \ref{table1}, the \textbf{dictionary} setting performs slightly worse than the \textbf{char1} setting, although it has lower TSVS and ALFL. This suggests that the vocabulary size and query length are not the only factors that affect the eventual accuracy. Perhaps the frequency of the tokens seen by the model during the pre-training task plays a role. It is likely that the model has encountered, during pre-training, single characters a far larger number of times than the words used in \textbf{dictionary} vocabulary. 

\section{Error Analysis}

We performed an error analysis on a sample of 100 randomly selected questions which produced an incorrect output. In the \textbf{original} setting, roughly 50\% errors were due to the presence of non-printable characters in the query (eg: \textasciicircum). We found that in the initial masked query, while we had replaced some non-printable characters in the pre-processing stage (eg: \{, \} ), we had not managed to replace the full set of non-printable characters. The original T5 paper mentions curly braces as one of the class of tokens that are not present in the pre-training corpus, however, a comprehensive list of the tokens that do not work with T5, or work with limited efficiency, is not available. In this scenario, it seems that a better approach is to replace the entire vocabulary with one that is entirely known to T5, for example, English words.  When comparing errors made by \textbf{original}, that were fixed by \textbf{dictionary} and \textbf{char1}, we observed that roughly 30\% of the cases were of variable placement, where the variable placeholders like \texttt{ent0}, \texttt{rel0} were found to be in the wrong order in the output query in the \textbf{original} setting. Rest of the corrections belonged to the category of syntax errors. This points to the finding that alternate vocabularies improve the ability of T5 to correctly produce logical forms from a semantic perspective.  

To analyse the effect of increasing complexity of vocabulary, we compare 100 randomly selected errors made by \textbf{char8} with \textbf{char2}. In both these settings, no character is non-printable, and the only errors are either syntax errors, variable placement errors, structural errors or intent errors. Out of the 100 questions, 90 were found to be correct in \textbf{char2} setting. In the remaining 90 in the \textbf{char8} setting, the highest proportion of errors belonged to syntax (where the query is malformed). The next most prominent class of errors belonged to variable placement, followed by structural errors (eg: two triples instead of three). The major takeaway from this analysis is that for \textbf{char2} there were no syntax errors, while in \textbf{char8} there are a significant number of such errors. 

\section{Conclusion}

In this work we carried out experiments with new output vocabularies, where we carefully substituted the original members of the vocabulary with the new ones. We found that when the original SPARQL vocabulary is replaced with words from an alternate vocabulary closer to the T5 tokenizer vocabulary, the model consistently perform better. 

As a contribution, we 
believe that our findings will enable researchers in the field of semantic parsing to deploy smaller models with a modified vocabulary and still find satisfactory performance. This would, in the longer term, lead to energy savings.

As future work, we would like to explore the behaviour of the same models in more depth using attention maps. Moreover, the significant shift in initial performance on changing vocabulary from \textbf{original} to \textbf{char} and 
\textbf{dictionary} demands further investigation. Similarly, the relatively lower performance of the \textbf{dictionary} setting when compared to \textbf{char1} setting, in spite of having lower tokenized vocabulary size (TSVS) needs to be investigated further. Perhaps sub-words which are seen more frequently during pre-training task of the LM perform better when substituted into the semantic parsing output vocabulary.

\section{Limitations}

We found that prefix tuning takes much longer to converge when compared to fine tuning, and for T5-Base, it takes around 10 days on a 48 GB GPU to complete tuning for a single setting in Table \ref{table1}. Due to limitation of resources and with an aim to save energy, we did not conduct experiments with larger models such as T5-Large, T5-XL etc. We also did not perform experiments with smaller splits of the same datasets, which could have given further insights on how model performance varies when training data size is less. 

\bibliography{anthology,custom}
\bibliographystyle{acl_natbib}

\appendix
\section{Samples}

\begin{table*}[t!]
\small
  \centering
  \begin{tabular}{|p{2cm}|p{13cm}|}
      \hline
      &GrailQA  \\
      \hline
    Question & Military airfield is the type for what airport ? \\
    \hline
    SPARQL  &
    \begin{verbatim}

SELECT DISTINCT ?x0  WHERE {
  ?x0 :type.object.type :aviation.airport .
  VALUES ?x1 { :m.0199qf }
  ?x0 :aviation.airport.airport_type ?x1 .
  FILTER ( ?x0 != ?x1  )
}

\end{verbatim} 
\\
    \hline
    Masked Query (\textbf{original} \\setting)   &  \begin{verbatim}
SELECT DISTINCT ?x0 WHERE OB 
  ?x0 :type.object.type rel0 . 
  VALUES ?x1 OB ent0 CB 
  ?x0 rel1 ?x1 . 
  FILTER ( ?x0 != ?x1 ) 
CB 
\end{verbatim}
    \\
    \hline
   dictionary &    
    \begin{verbatim}
banana compound boy nation rain 
  boy catastrophe elementary flower 
  teeth today rain jacket case 
  boy fog today flower 
  duck folk boy chart today concede 
case 
\end{verbatim}
    \\
     \hline
 char1 &    
    \begin{verbatim}
- 1 A Y $
  A : O % 
  L J $ G S 
  A | J % 
  0 M A + J X 
S
\end{verbatim}
    \\
     \hline
    char2 &    
    \begin{verbatim}
UY SJ 0X 6L VZ 
  0X 5G JO SE 
  5Z QB VZ QJ 8O 
  0X FT QB SE 
  RU 2K 0X WY QB I5 
8O 
\end{verbatim}
    \\
     \hline
    char4 &    
    \begin{verbatim}
53IY 3UQZ JKMQ CEK2 5DZV 
  JKMQ KRDN 1G8E ZC5C 
  5ILL 3JBD 5DZV X5XB YMG5 
  JKMQ ZVGC 3JBD ZC5C 
  87O2 DE3Z JKMQ TU76 3JBD 049K 
YMG5
\end{verbatim}
    \\
\hline
    char8 &    
    \begin{verbatim}
WDEUTG57 L741BHJP ORWDXYPH 6L05N8AS ZLZXSARH 
  ORWDXYPH K4GR9TPQ 797G3PGO V13Y1EFE 
  PQMAIPQ4 MLN1V72G ZLZXSARH KPHC8I2N WG0XRTYG 
  ORWDXYPH ZF82YUH8 MLN1V72G V13Y1EFE 
  41O2LA2M F1SANW03 ORWDXYPH 4R26K1BW MLN1V72G TD9BSKSN 
WG0XRTYG
\end{verbatim}
   \\
  
   \hline
 \end{tabular}
 
\caption{An example of a question from GrailQA, with the corresponding SPARQL query, and how they look once new vocabularies are substituted.}
\label{tablevocabs}
\end{table*}




\end{document}